\documentclass{article}
\usepackage[latin1]{inputenc}
\usepackage{amsmath}
\usepackage{amsfonts}
\usepackage{amssymb}
\usepackage{makecell}
\usepackage{booktabs}
\usepackage{graphicx}

\usepackage[tight,footnotesize]{subfigure}

\usepackage{color}

\usepackage{mathtools}

\usepackage{color}

\usepackage{units}

\author{Claudia Carpineti, Vincenzo Lomonaco, Luca Bedogni, \\Marco Di Felice, Luciano Bononi\\
Department of Computer Science and Engineering, \\University of Bologna, Italy\\
Email: \{claudia.carpineti, vincenzo.lomonaco, \\luca.bedogni4, marco.difelice3, \\luciano.bononi\}@unibo.it}

\date{}

\newcommand{\ls}[1]
  {\dimen0=\fontdimen6\the\font
   \lineskip=#1\dimen0
   \advance\lineskip.5\fontdimen5\the\font
   \advance\lineskip-\dimen0
   \lineskiplimit=.9\lineskip
   \baselineskip=\lineskip
   \advance\baselineskip\dimen0
   \normallineskip\lineskip
   \normallineskiplimit\lineskiplimit
   \normalbaselineskip\baselineskip
   \ignorespaces
}

\title{Custom Dual Transportation Mode Detection by Smartphone Devices Exploiting Sensor Diversity}

\begin{document}
\maketitle

\begin{abstract}
Making applications aware of the mobility experienced by the user can open the door to a wide range of novel services in different use-cases, from smart parking to vehicular traffic monitoring. In the literature, there are many different studies demonstrating the theoretical possibility of performing Transportation Mode Detection (TMD) by mining smartphones embedded sensors data. However, very few of them provide details on the benchmarking process and on how to implement the detection process in practice. In this study, we provide guidelines and fundamental results that can be useful for both researcher and practitioners aiming at implementing a working TMD system. These guidelines consist of three main contributions. First, we detail the construction of a training dataset, gathered by heterogeneous users and including five different transportation modes; the dataset is made available to the research community as reference benchmark. Second, we provide an in-depth analysis of the sensor-relevance for the case of Dual TDM, which is required by most of mobility-aware applications. Third, we investigate the possibility to perform TMD of unknown users/instances not present in the training set and we compare with state-of-the-art Android APIs for activity recognition.
\end{abstract}

\section{Introduction}
The term ``\emph{context-aware computing}'' was coined first in 1994, and denoted as ``\emph{the possibility to  exploit the changing environment with a new class of applications that are aware of the context in which they are run}'' \cite{Schilit1994}. In recent years, such possibility has become more and more concrete thanks to the pervasive diffusion of smartphone devices enabling anytime/anywhere computing and Internet connectivity. Moreover, modern smartphone devices are typically equipped with a wide range of embedded sensors, through which it is possible to sense the surrounding environment, and also to detect the user's location and the activity being performed. \\ 
In this work, we focus on a specific problem of the Human Activity Recognition (HAR) discipline \cite{Lara2013}, i.e. on how to infer the transportation mode experienced by the user, on the basis of the smartphone sensors. Many popular mobile APPlications (APP) include functionalities of automatic Transportation Mode Detection (TMD) \cite{Bedogni2016}, for instance related to the execution of automated actions in response of detected locations and mobility-aware events; we cite, among other, the popular If This Than That (IFTTT) framework \cite{ifttt}. In large-scale urban scenarios, TMD techniques are also employed to enable the seamless gathering of mobility traces on a voluntary basis, avoiding the need of external sensing infrastructures, and at the same time minimizing the annoyance for the users \cite{Su2016}.  

At present, TMD is performed by two different techniques, namely GPS based \cite{Reddy2010} or sensor based \cite{Bedogni2016}. Other studies complement the devices data with external information about the scenario (e.g. the transportation system map) \cite{Stenneth2011}. Moreover, there exists a vast literature on the design and analysis of pattern matching techniques aimed at maximizing the accuracy of the detection process  \cite{Reddy2010}\cite{Bhattacharya2016}\cite{Su2016}. Among the non-research related initiatives, it is worth remarking that the Android operating system, since version 4.0,  offers convenient APIs to the APPs that can be informed about the current transportation mode, although the set of recognized actions is limited to four classes (\texttt{WALKING}, \texttt{STILL}, \texttt{VEHICLE}, \texttt{BIKE}). 

At the same time, the existing studies on TMD techniques still suffer of two main limitations. First, their results are hard to compare due to the diversity of the data/sensors used, and, in some cases, results are not generalizable due to the limited number of users involved in the experiments. At the best of our knowledge, no dataset has been released for public usage and validation by the research community which can address these shortcomings. Second, all these works assume that the recognition must occur among all the available transportation modes in the training set, which is  clearly the hardest case for the classifier. However, in several realistic use-cases the APP must only distinguish between two classes; this is the case, for instance, of smart parking systems that must distinguish between \texttt{WALKING} and \texttt{VEHICLE} modes, in order to trigger the proper actions (e.g. send a notification about a new free/busy spot) \cite{Krieg2016}\cite{Salpietro2016}. In this paper, we provide results that can serve as guidelines for researchers and  developers interested in practically designing and implementing Dual TMD systems. More in details, we provide three main contributions:
\begin{itemize}
	\item We build a new dataset called \textsc{US-TMD} (\textit{Unconstrained Sensors Transportation Mode Dataset}) comprising 13 users, with nearly 32 hours of total recordings, and we make it publicly available to the research community. We also detail the methodology adopted for the dataset population and the preliminary data filtering and pre-processing.
	\item We perform a detailed study of Dual TMD over the dataset, and we report the results about what sensor is relevant to which transportation mode.
	\item We investigate the relationship between user-aware and user-agnostic training process, and we show by experimental results that a model trained over the proposed dataset is able to recognize actions from unknown users with high accuracy.
\end{itemize}
Finally, we integrate our results with a state-of-the-art comparison with the Google Activity Recognition API which has been collected during the data acquisition process.\\ 
The rest of this paper is organized as follows: in Section \ref{sec:rel} we present the related work from literature for Transportation Mode Detection; Section \ref{sec:dataset} introduces our dataset, and the acquisition/pre-process methodology we used; Section \ref{sec:evaluation} presents the main analysis performed on the dataset (class-vs-class tests, leave-one-out tests and Google API comparison) and Section \ref{sec:conclusion} concludes the work, and discusses future works on the topic.

\section{Related Work}
\label{sec:rel}
Research on context aware computing has been revamped in the last few years, mainly thanks to the pervasive diffusion of modern smartphones which are equipped with a wide set of embedded sensors. We identify  three different methods for assessing the user transportation mode: GPS-based, sensor based, or external-source based.  The former is quite popular, mainly due to the wide availability of GPS in modern smartphones and for its accuracy in discriminating between motorized vehicles and pedestrian activities \cite{Reddy2010}. However, it suffers from heavy battery consumption and scarce accuracy in indoor environments or urban canyons \cite{Su2016}, due to fading and multipath signals which lower the GPS accuracy. Moreover, it is unable to correctly classify transportation modes with similar speeds \cite{Bedogni2016}. 

Sensor-based approaches are often based on Machine Learning (ML) techniques and on training set of classified instances. For location-awareness, the magnetometer showed good performance in locating the user through fingerprinting maps \cite{Kang2015a}, while the barometer has been used for locating the altitude from the ground level \cite{Bedogni2016a}. For TMD, the most used sensor is certainly the accelerometer, which provides the best trade-off between the accuracy of the activity recognized and the energy consumption \cite{Bedogni2016} \cite{Kwapisz2010} \cite{Hemminki2013}. External-source based approaches enhance the robustness of TMD algorithms by extra-source information that are relative to the current user location. In \cite{Stenneth2011}, the authors consider external transportation network data such as real time bus locations in addition to the GPS data, in order to distinguish among different motorized transportation modes with similar speed readings. 
In \cite{Bloch2015}, the authors introduce a two step detection approach, where mobility events are detected by analyzing the cellular signal strength, and then -in case of real location change- the transportation mode is classified by means of the GPS and the accelerometer.  We also highlight that there is a wide literature on the algorithms for TMD, including both  the application/evaluation of well-known ML techniques (e.g. Random Forest is largely used in  \cite{Bedogni2016}\cite{Reddy2010}\cite{Stenneth2011}), and the design of novel recognition schemes \cite{Bhattacharya2016}. \\
However, at the best of our knowledge, no prior study addressed the problem of Dual TDM (i.e. the case when we must discriminate with high accuracy between two classes only). Moreover, few studies provided extensive analysis about the sensor relevance when classifying a specific transportation mode. The most similar works are \cite{Bao2004} and \cite{Casale2011}. In \cite{Bao2004}, the authors presented a study on activity recognition through sensor values, although people involved in the experiments have to carry several sensors at fixed positions of the body. Authors of \cite{Casale2011} present a study on the classification of human daily activities, such as \texttt{WALKING} and \texttt{WORKING}. They also present the results of the RF classifier about the feature importance. However, similarly to \cite{Bao2004}, they make assumption about the location and orientation of the devices.
Finally, concerning freely accessible datasets in this context, the GeoLife dataset \cite{zheng2010geolife} constitutes a valid resource for many transportation modes but it collects only GPS data and with a low recording frequency (1-5 sec.). Hence, being able to thoroughly compare different methods for TMD is currently impossible due to the lack of common and shared benchmarks with recording sessions collecting both GPS-based and sensor-based measures.

\section{US-TMD Dataset}
\label{sec:dataset}

In light of the lack in the literature of a common benchmark for TMD, we have collected a large set of measurements belonging to different users and through a simple Android APP. 
the \textsc{US-TMD} is built from people of different gender, age, and occupation. Moreover, we do not impose any restriction on the use of the application, hence every user records the data performing the action as she/he is used to, in order to assess real world conditions.
We openly release the dataset, so that other researchers can benefit from it for further improvements and research reproducibility. 

Sensors data are collected from thirteen volunteer subjects, ten male, and three female. Table \ref{tab:DSdevice} summarizes the data collected by users by looking at five dimensions: gender, age, occupation, device model, and Android version installed while they collected data. The set of classes considered is composed of \texttt{WALKING}, \texttt{CAR}, \texttt{STILL}, \texttt{TRAIN} and \texttt{BUS}. This follows common practices in literature \cite{Bedogni2016} \cite{Reddy2010}.

\begin{table}[t!]
\centering
\setlength\tabcolsep{3.4pt}
\def\arraystretch{2}
\begin{tabular}{|c|c|c|c|c|c|}
	\hline
	 ID & \textbf{Sex} & \textbf{Age} & \textbf{Occupation} & \textbf{Device} & \textbf{Android Version} \\
 	\hline			
  	$U_{1}$&M&30&student&LG G2& 5.0.2 \\
  	\hline
  	$U_{2}$&F&27&student& \makecell{Sony XPERIA Z3\\ Compact D5803} & 6.0.1 \\
  	\hline
  	$U_{3}$&M&30&student&Nexus 5& 7.0\\
  	\hline
  	$U_{4}$&M&36&office worker&Huawei Honor 5X& 6.0.1\\
  	\hline
  	$U_{5}$&M&36&stage director&Huawei P8  Lite& 6.0.1 \\
  	\hline
  	$U_{6}$&M&27&researcher&\makecell{Samsung galaxy\\ s3 neo}&4.4.2  \\
  	\hline
  	$U_{7}$&M&32&cameramen&Samsung S7&6.0.1 \\
  	\hline
  	$U_{8}$&F&32&bartender&Huawei Tag-l01& 5.1\\
  	\hline
  	$U_{9}$&F&24&student&Motorola Moto G& 5.1\\
  	\hline
  	$U_{10}$&M&22&student&Huawei P9 & 7.0\\
  	\hline
  	$U_{11}$&F&31&office worker&Nexus 5&7.0\\
  	\hline
  	$U_{12}$&M&31&researcher&Samsung Galaxy S6&6.0.1\\
  	\hline
  	$U_{13}$&M&60&retired&Nexus 5&7.0\\
  	\hline
\end{tabular}\\~\\
\caption{Dataset variability in terms of users' age, sex, occupation, device model and Android version.}
 \label{tab:DSdevice}
\end{table}

 \begin{figure*}[t]
    \centering
    \subfigure[Sensors row data]{
        \includegraphics[width=0.45\textwidth]{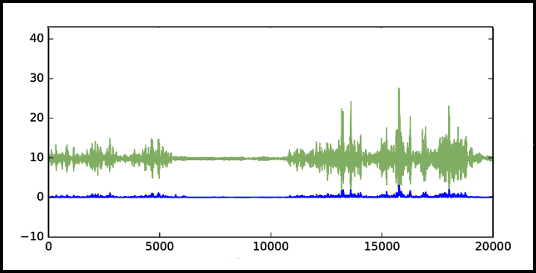}
        \label{fig:carMagnitude}
        }
    \subfigure[Windows partitioning]{
        \includegraphics[width=0.45\textwidth]{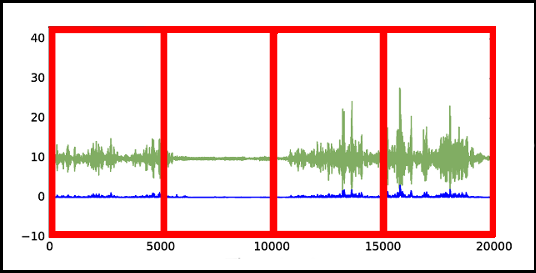}
        \label{fig:walkingMagnitude}
		}	
	\subfigure[Features extraction]{
        \includegraphics[width=0.45\textwidth]{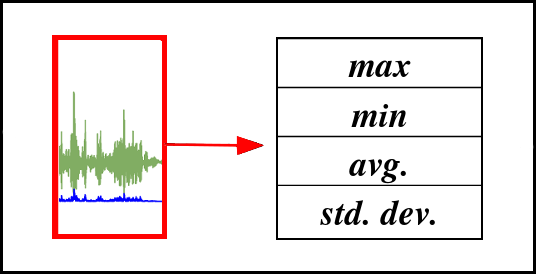}
        \label{fig:trainMagnitude}
        }
    \subfigure[New data predictions]{
        \includegraphics[width=0.45\textwidth]{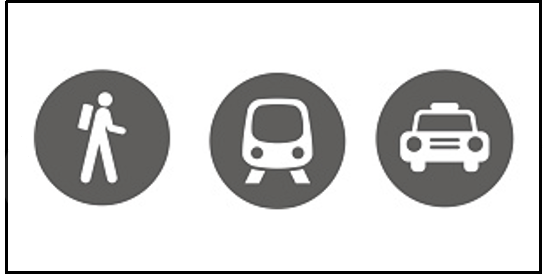}
        \label{fig:stillMagnitude}
        }
    \caption{\textit{The Main process behind the construction of a TMD system}. a) Firstly, row data belonging to different sensors are collected through the APP while the subjects are freely performing their activities. b) the time series for each sensor are then split in time windows of a fixed size. c) For each window, standard and robust numerical features are extracted (such as min, max, mean and standard deviation) the ML model is trained on the extracted features for each sensor. Finally, in d) activities from new streaming sensors data can be predicted.}\label{fig:magnitude}
\end{figure*}


The raw sensor data collection is performed by the application which registers each sensor event with a maximum frequency of 20 Hz. Events occurs every time a sensor detects a change in the parameters it is measuring, providing four different information, such as the name of the sensor, the timestamp, the accuracy and the raw data from the sensor. The length and content of the raw sensor data depend on the type of sensor, which we record as
$$<timestamp, sensor_{i}, sensorOutput_{i}>$$
We save each of these measurement on CSV file on the Android device, which is then uploaded to our server for the data processing. 

In total, our dataset is composed of 226 labeled csv files representing the same number of activities corresponding to more than 31 hours of data: 26\% of data is annotated as walking, 25\% as driving a car, 24\% as standing still, 20\% as being on a train, and 5\% as being on a bus. More detailed information about the dataset can be found in its website\footnote{http://cs.unibo.it/projects/us-tm2017}, where we also report the total recorded time for each user in the dataset.
\begin{table}[h!]
\centering
\def\arraystretch{2}
\setlength\tabcolsep{2.5pt}
\begin{tabular}{|c|c|c|c|c|c|c|}
	\hline
	 \textbf{Bus} & \textbf{Car} & \textbf{Still} & \textbf{Train} & \textbf{Walking} & Total \\
 	\hline			
 	\textbf{01:44:35}& \textbf{07:53:50}&  \textbf{07:29:35}&  \textbf{06:20:25}&  \textbf{08:20:25}&\textbf{31:48:50} \\
 	\hline
\end{tabular}\\~\\
\caption{Dimension of dataset in terms of recording times for class.}
 \label{tab:DStime}
\end{table}


Although some sensors may not be useful for the purpose of TMD, we collect data from each sensor available in the mobile devices considered, and offer them in our public dataset. Indeed, since we aim to provide a common benchmark for TMD we leave the selection of the sensors data to use open to new judgments (depending on the classes to distinguish and to the trade-off between accuracy and resource consumption) but still providing comparability with a common and comprehensive test set which has been missing so far.
The analysis we perform in Section \ref{sec:evaluation} better details which of these are more important than others in classifying an action.

As described in In Figure \ref{fig:magnitude}, the first step for creating the features which will be used to construct the machine learning model is to partition the dataset in time windows. The size of the time window depends on the types of actions to be recognized, and in this work we set it to \unit[5]{seconds}, which is a common value for similar task also for related works in literature \cite{Bedogni2016}. If the adopted time of the sliding window is too short, the window data may not have covered the information of a complete action. On the other hand, if the width of the sliding window is too long, it will not only make the data sophisticated, but also increase the amount of calculation.

In the study presented in this paper we aggregate data in 5 second windows, but we also offer to the research community the raw 20-Hz measurements so anyone can clusterize the data as needed. Also, windows of 5 seconds have already been used in literature \cite{Bedogni2016}, in which the authors also studied the benefits of overlapping windows.
We compute for each window four features based on the multiple raw sensor readings, the maximum (\texttt{max}), the minimum (\texttt{min}), the mean (\texttt{mean}) and the standard deviation (\texttt{std}). Therefore, for each window we end up with a total of $\Lambda \cdot 4$ features, where $\Lambda$ is the total number of sensors which reported a value in the given time window. Missing values are then filled with average values on the training set for each sensor.

%


\section{Evaluation}
\label{sec:evaluation}

In this section we present performance evaluation regarding classification on our dataset. Although the dataset offers many sensors, we have decided to perform an analysis on a subset of them, motivated by the following rationale. Basically, some of the sensor introduce noise, as they are not representative of the transportation mode but of the location in which the data has been recorded. Therefore, since the ML algorithm may wrongly leverage some biases contained in this sensors we have decided to excluded them from our analysis. The excluded sensors are: \texttt{light}, \texttt{pression}, \texttt{magnetic field}, \texttt{gravity} and \texttt{proximity}.

From the remaining sensors, we have created three evaluation datasets, namely $D^1$, $D^2$ and $D^3$. $D^1$ is composed by the \texttt{accelerometer}, \texttt{gyroscope} and \texttt{sound}; $D^2$ contains all the others but the \texttt{speed}, which is added in $D^3$. 

For each $D^i$, we build four models with four different classification algorithms: Decision Trees (DT), Random Forest (RF), Support Vector Machines (SVM), and Neural Network (NN). Since NN and SVM require precise parameters characterization (\cite{hsu2003practical}, \cite{bengio2012practical}), we perform a 10-fold cross validation to find the best parameters, and select those for the rest of our analysis.
$D^1$ is built taking three low-battery-consumption sensors, among those typically available in smartphones and which gave the best single-sensors accuracies. Increasing the number of sensors used, as we did in $D^2$, the accuracy can slightly increase, while keeping the GPS, the most consuming sensor, out of our subset. $D^3$ reports instead the use of all the sensors, including the GPS. 
Before proceeding with the training of the models a class representation balancing of the training set may be needed. Indeed, since the bus has a much lower representation it may be convenient for the model to favor other motorized classes in the prediction. In this work we decided to truncate all the classes representation to 1:44:35 hours (which is the lower recorded time for the bus class) maintaining the proportional contribution of each user.

Detailed values for overall accuracy in all different dataset and for all the algorithms are reported in Table \ref{tab:overAllAccuracy}.

\begin{table}[h!]
\def\arraystretch{2}
\setlength\tabcolsep{6pt}
\begin{tabular}{|c|c|c|c|}
\hline
\textbf{Algorithm}&\makecell{\textbf{Accuracy on}\\\textbf{$D^1$}}&\makecell{\textbf{Accuracy on}\\\textbf{$D^2$}}&\makecell{\textbf{Accuracy of}\\\textbf{$D^3$}}\\
\hline
Decision Tree (DT)& 76\% &78\%&86\%\\
\hline
Random Forest (RF)& 81\%&89\%&93\%\\
\hline
\makecell{Support Vector\\Machine (SVM)} &76\%&86\%&90\%\\ 
\hline
Neural Network (NN) & 76\%&87\%&91\%\\ 
\hline
\end{tabular}
\caption{Overall accuracy with all four classification algorithm.}
\label{tab:overAllAccuracy}
\end{table}

However, typically context-aware applications do not need to recognize all the classes as we did, since they generally need only a subset of them. Therefore, the next analysis we perform is devoted to the class-to-class classification, in which we perform a similar analysis but considering only two classes to discriminate between.

\subsection{Class-to-class classification}
\begin{figure}[t!]
	\includegraphics[width=0.95\textwidth]{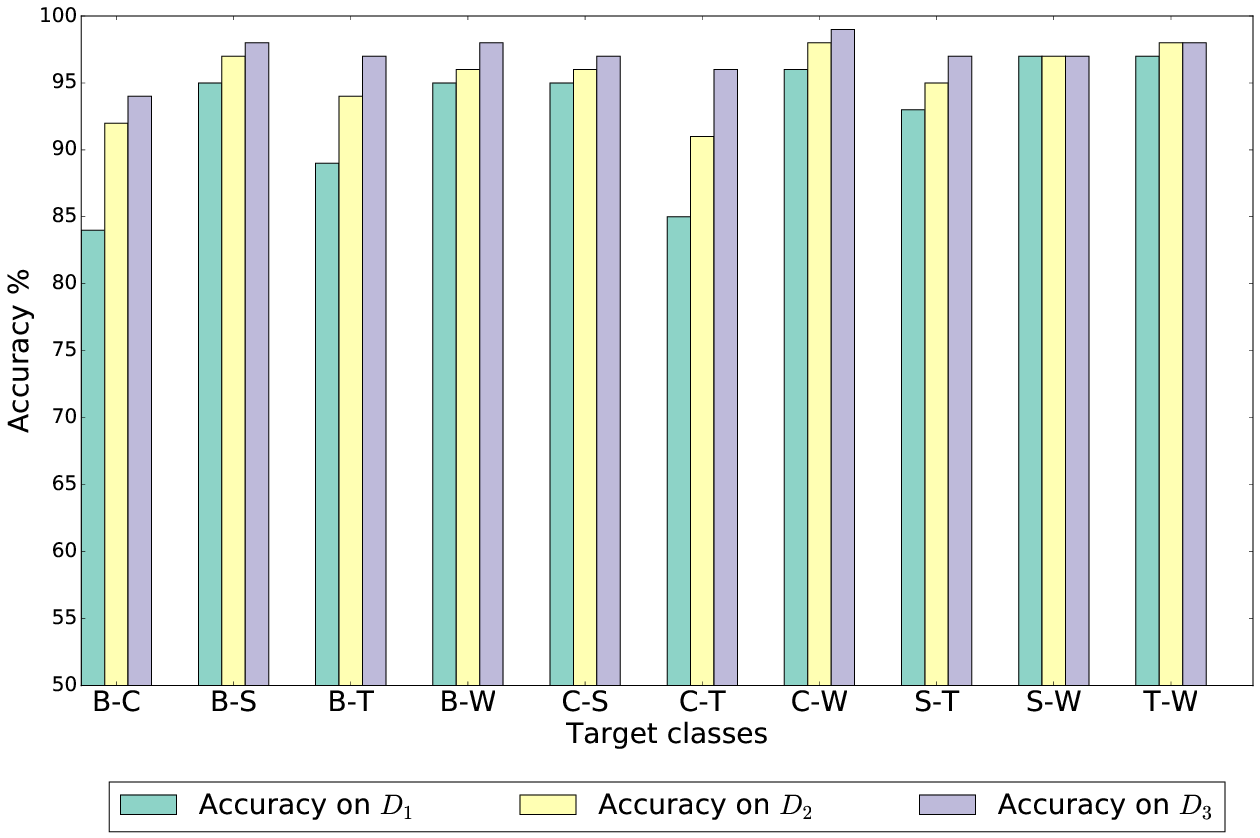}
	\caption{\textit{Class-vs-class accuracy on the test set with respect to the three different sensors sets $D^1$, $D^2$ and $D^3$}. Letters B, C, S, W, T stand for \textsc{B}us, \textsc{C}ar, \textsc{S}till, \textsc{W}alking and \textsc{T}rain respectively. Better viewed on colors.}
	\label{fig:classtoclass}
	\vspace{-0.5cm}
\end{figure}

Obviously, reducing the amount of classes to be classified raises without any exception the accuracy of any classification algorithm. However, in this paper we report only the results from the RF, since it has shown the best results also for this analysis. 
Figure \ref{fig:classtoclass} reports the accuracy when classifying all the possible couples of classes. At first, it is straightforward to note how some couples are more challenging than others, especially for $D^1$ and $D^2$. For instance, \texttt{\{Bus,Car\}}, \texttt{\{Bus,Train\}} and \texttt{\{Car,Train\}} highlight a considerable increase in accuracy when switching from $D^1$ to $D^2$ and eventually to $D^3$, hence confirming the importance of the speed feature for motorized classes. On the other hand, recognizing couples of activities in which one is \texttt{Walking} is easier even for $D^1$. Hence, for those tasks the importance of having the features obtained from the speed is lower.

\subsection{Sensor importance}
\begin{figure}
	\centering
	\includegraphics[width=0.95\textwidth]{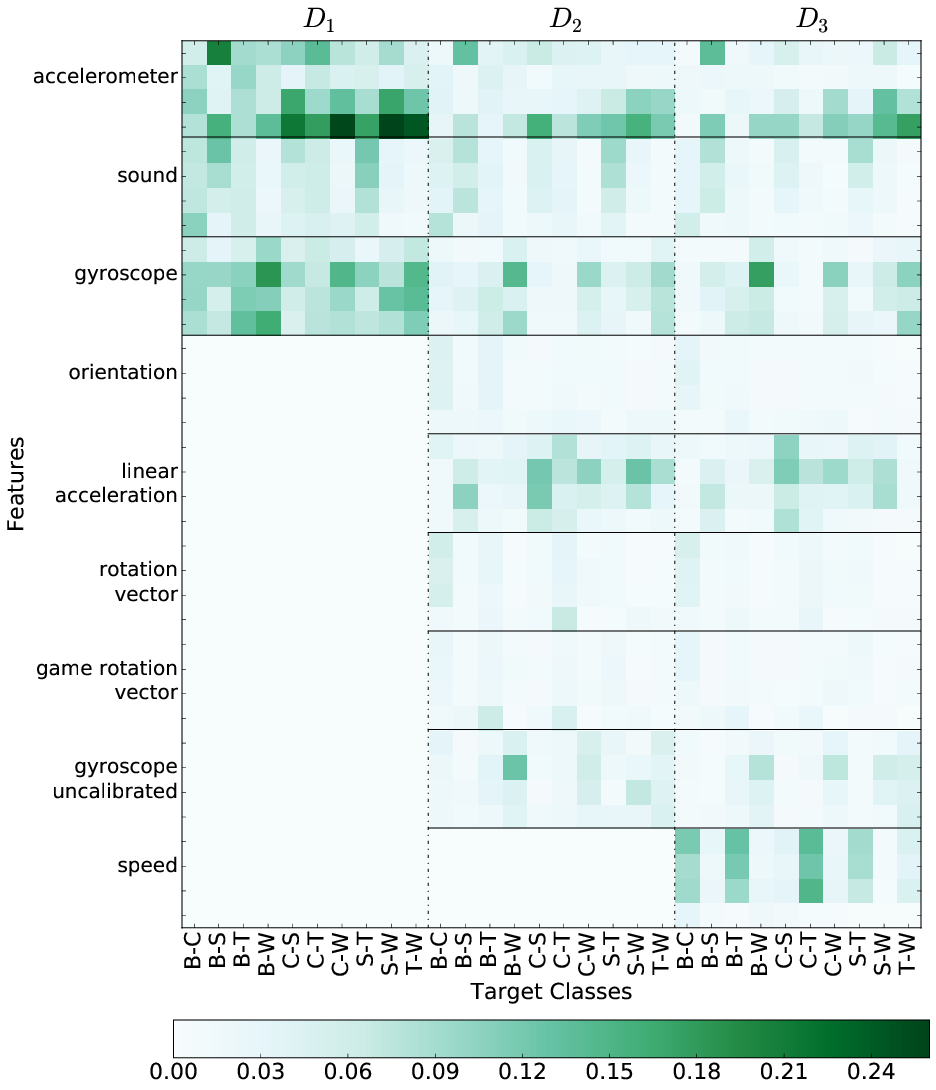}
	\caption{\textit{Sensor importance on $D^1$, $D^2$ and $D^3$ (from left to right) for each class-vs-class setting}. High intensity color stands for more important sensors. Better viewed on colors.}
	\label{fig:sensorimportance}
	
\end{figure}

To analyze deeper the results from the previous sections, and to understand how and why some classes are easier to be recognized compared to others, we now perform an analysis on the importance of each sensor in classifying a specific set of 2 classes.
We show the results from this analysis in Figure \ref{fig:sensorimportance}, where we plot the 10 couples of classes on the x axis, for $D^1$, $D^2$ and $D^3$, and on the y axis we show all the possible features. Clearly, for datasets in which such feature is not available, the corresponding square will be left blank. For $D^1$, the leftmost column of Figure \ref{fig:sensorimportance}, we can see how features obtained from the accelerometer are by far the most important in recognizing the classes, followed by the gyroscope. Features from the sound have instead less importance. Moving to $D^2$, the gyroscope loses importance, while the accelerometer (especially the standard deviation) keeps its importance high. Moreover, also the linear acceleration, obtained from the raw values of the accelerometer, is considered important by the model. Finally, $D^3$ still considers the accelerometer and the linear acceleration as important features, but the just introduced speed is considered important as well, except for the standard deviation which instead have low relevance for distinguishing the different classes. This somehow confirms the results of \cite{Bedogni2016} considering models with only the GPS. In addition, it is evident how the speed is considered important by the model only for motorized classes.

An interesting aspect emerges, related to the fact that whenever one of the classes to be classified is \texttt{WALKING}, features obtained from the accelerometer and the gyroscope are identified as the most relevant and representatives of the movement. 



Consequently, a model the aim of which is to figure out if a user is \texttt{WALKING} can be easily based on these sensors. On the other hand, when one of the classes to be classified is \texttt{STILL}, also the sound can be helpful. 

\subsection{Leave one out Analysis}
In this section we detail the last analysis we performed, which is called Leave one out. In this particular analysis, we aim at understanding whether crowd based models, based on raw data coming from a number of individuals, can be used to classify the transportation mode of a user not taking part in the training phase. This would be important, for instance, to provide general models which can be used by the final user, which might possibly refine it by adding data from her/his own measurements.


\begin{figure}
	\centering
	\includegraphics[width=0.95\textwidth]{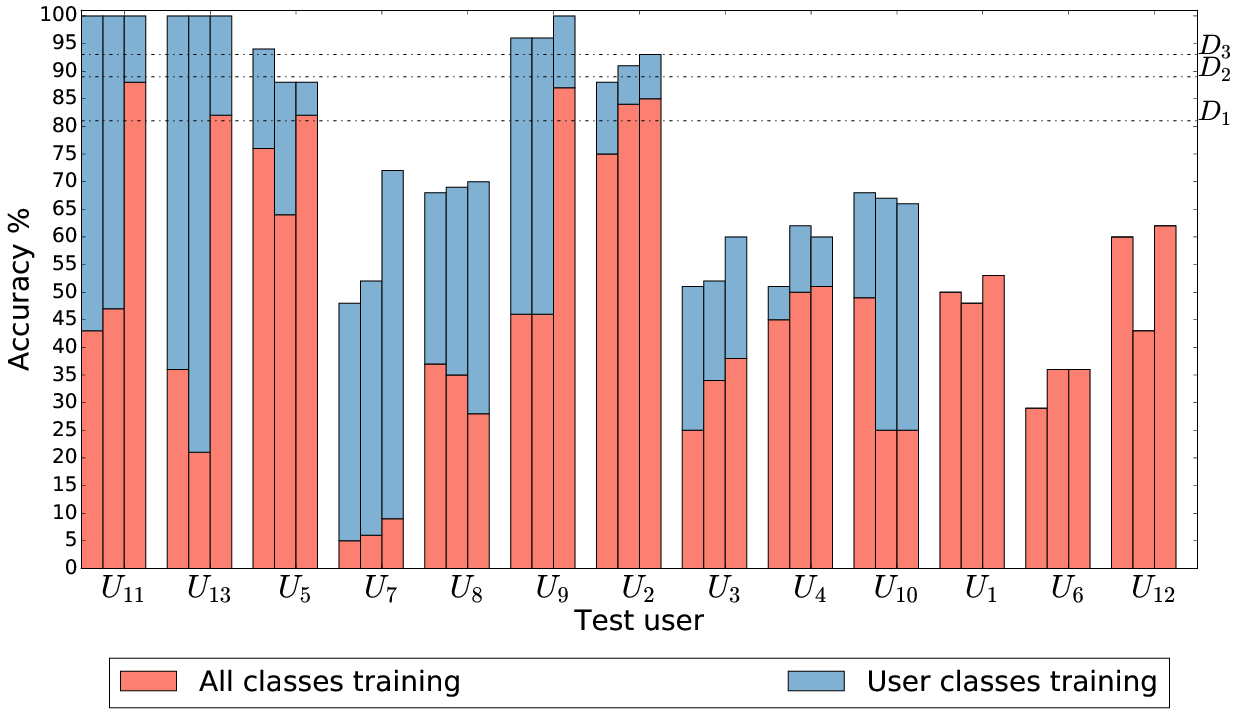}
	\caption{\textit{Leave One Out Test}. For each user, the accuracy levels of the model trained on all the other users and on the three sensors set $D^1$, $D^2$ and $D^3$ (from left to right) are reported. Blue bars stand for the accuracy results considering only the class of interest for that particular user, red for the accuracy considering all the classes. Test set results with all the users included and for each sensor set are reported as dotted lines for comparison. Users are ordered from left to right by number of classes to distinguish. Better viewed on colors.}
	\label{fig:leaveoneout}
\end{figure}

Figure \ref{fig:leaveoneout} reports the results from this analysis. 
For each user, red bar represents a scenario in which the training is performed on all the 5 classes, and the recognition only on the classes available for such user. The blue bar represents instead a case in which the training is obtained only from data relevant to the classes to be recognized for the given user (although not including any data from her/him). The three bars for each user are obtained by using $D^1$, $D^2$ and $D^3$, respectively. The first comment we can make is that if in the training there is no data of a particular user, the accuracy of the model is quite low, well below the average presented in Table \ref{tab:overAllAccuracy}. When instead the training is performed only on the classes relevant for that user, the accuracy increases for each one. 

Interestingly, there are some users which appear to be ``easier'' to be recognized than others, keeping their accuracy high regardless of $D^1$, $D^2$ or $D^3$. Others instead achieve higher accuracies only when using a model training on their specific classes, which thus reduces the possible errors coming from wrong classification.

\subsection{Google Activity Recognition}
In this section we also provide performance evaluation, on the same dataset, for the Google Activity Recognition API, which offer a convenient method for Android developers to obtain the transportation mode of the device\footnote{https://developers.google.com/android/reference/com/google/android/\\gms/location/ActivityRecognition}.
Basically, an Android APP can register to such events, and be notified by the operating system whenever a new transportation mode is detected. The Transportation Mode classes that the service recognizes are \texttt{IN\_VEHICLE}, \texttt{ON\_BICYCLE}, \texttt{ON\_FOOT} (referring to a user walking or running), \texttt{RUNNING}, \texttt{WALKING}, \texttt{STILL}, \texttt{TILTING}, and \texttt{UNKNOWN}.

Even tough we can not directly compare the quality of our algorithms with respect to the Google API (since the underneath algorithm has been trained on a different training set and with unknown methodologies), we can still analyze the results of the API on our test set.
Since the set of classes differs between the ones that Google aims to recognize and the ones we used in this paper, at first we perform a mapping between the Google classes and ours. The \texttt{RUNNING}, \texttt{WALKING} and \texttt{ON\_FOOT} are all three mapped on our \texttt{WALKING} class, \texttt{VEHICLE} can refer to \texttt{CAR}, \texttt{TRAIN} and \texttt{BUS}, and \texttt{STILL} is kept as it is.

\begin{figure}
	\centering
	\includegraphics[width=0.95\textwidth]{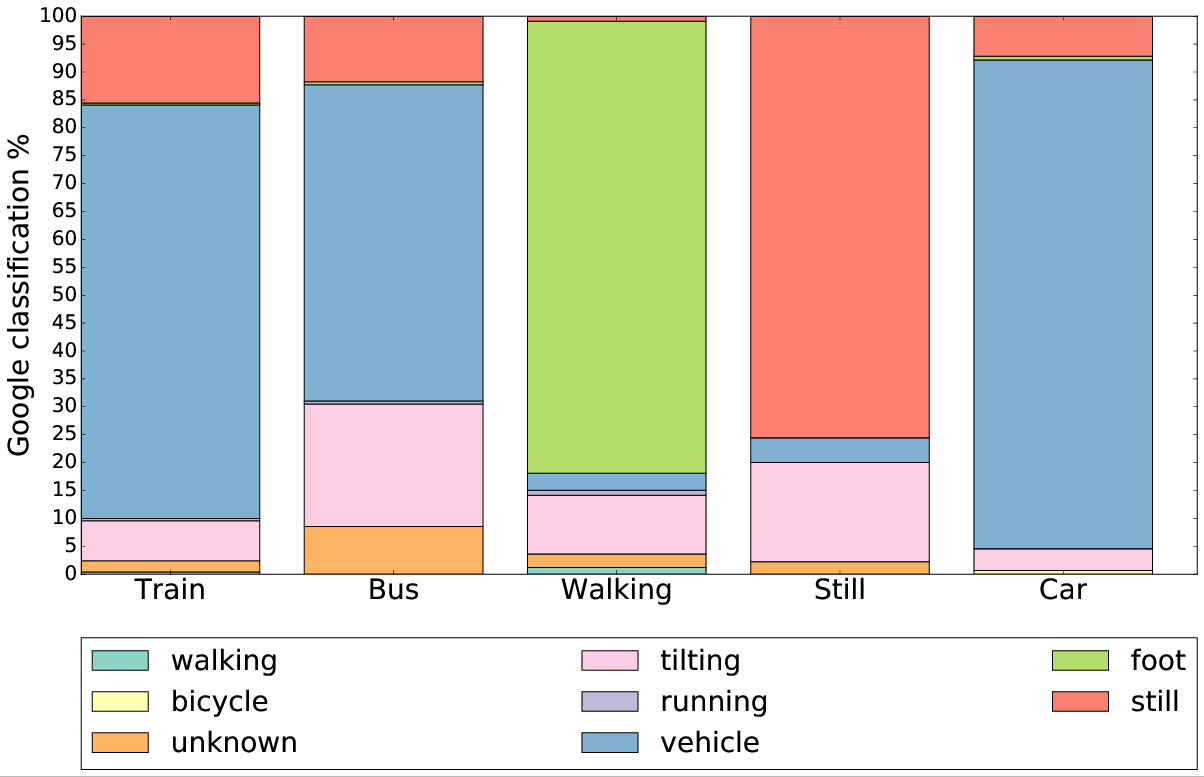}
	\caption{\textit{Google Activity Recognition API classification results}. For each Transportation mode on the x axis, percentage of inferred classes by the Google AR API are reported. We report as unknown also the cases for the test examples for which no response from the API is provided. }
	\label{fig:google}
\end{figure}

Figure \ref{fig:google} shows the classification results we detected for the Google Recognition API on our test set. At first, we note that out of the 22904 5-seconds time windows we have in the dataset, Google Activity Recognition API only classified 698 out of them, which represents roughly the \unit[3]{\%}. The \texttt{TRAIN} activity is often classified as \texttt{STILL}, while the \texttt{BUS} is often misclassified as \texttt{ON\_FOOT} or \texttt{RUNNING}. \texttt{WALKING} is recognized better than other classes, as well as \texttt{STILL} and \texttt{CAR}.
Clearly, the Google API model has to embrace a wider set of users and is much more conservative in its predictions (predicting unknown sometimes, since it is in an open-set scenario compared to ours). However, the much lower number of time windows classified might still constitute an issue for certain applications. A possible improvement would be to select the classes to be recognized by the APP and build a custom (eventually dual) TMD system as detailed in this work, so that the reduced uncertainty between the classes would lead to a sensible better accuracy with a constant prediction rate.

\section{Conclusion}
\label{sec:conclusion}
Context aware computing has risen to unprecedented levels, thanks to the proliferation of smart mobile devices. In particular, Transportation Mode is often considered as a valuable information for context aware applications, who can better exploit the context by knowing the mobility of the user. In this paper, we have presented three novel contributions: \textit{(i)} we have provided an open dataset, currently not available in the literature, that would help researcher to better study solutions which exploit Transportation Mode information; \textit{(ii)} A class-vs-class accuracy, highlighting that for specific applications it is better to limit the number of classes to the needed ones, rather than leveraging on tools which recognize uninteresting classes for a given service; \textit{(iii)} results on the possibility to use models trained on a crowd, which are later used to classify the Transportation Mode of unknown users.

Our results indicate that custom Transportation Mode Detection algorithms, tailored to the need of the application, always outperform those that recognize a wider set of classes. We have also shown that for some specific actions, a reduced set of sensors might still provide good classification accuracy, limiting also the battery consumption, a crucial aspect in mobile devices.

Future work on this topic is related to new approaches for improving the model transferability from one group of users to other individuals and the ability to incrementally update the model for improving the classification accuracy. Moreover, we foresee extending the open dataset, by including data from more users for a better model generalization.

\bibliographystyle{plain}
\bibliography{library.bib}

\begin{thebibliography}{10}

\bibitem{ifttt}
{IFTTT project - https://ifttt.com}.

\bibitem{Bao2004}
Ling Bao and Stephen~S Intille.
\newblock {Activity Recognition from User-Annotated Acceleration Data}.
\newblock {\em Pervasive Computing}, 3001:1--17, 2004.

\bibitem{Bedogni2016}
Luca Bedogni, Marco Di??Felice, and Luciano Bononi.
\newblock {Context-aware Android applications through transportation mode
  detection techniques}.
\newblock {\em Wireless Communications and Mobile Computing},
  16(16):2523--2541, 2016.

\bibitem{Bedogni2016a}
Luca Bedogni, Fabio Franzoso, and Luciano Bononi.
\newblock {A Self-Adapting Algorithm Based on Atmospheric Pressure to Localize
  Indoor Devices}.
\newblock In {\em 2016 IEEE Global Communications Conference (GLOBECOM)}, pages
  1--6. IEEE, dec 2016.

\bibitem{bengio2012practical}
Yoshua Bengio.
\newblock Practical recommendations for gradient-based training of deep
  architectures.
\newblock In {\em Neural networks: Tricks of the trade}, pages 437--478.
  Springer, 2012.

\bibitem{Bhattacharya2016}
Sourav Bhattacharya and Nicholas~D. Lane.
\newblock {From smart to deep: Robust activity recognition on smartwatches
  using deep learning}.
\newblock In {\em 2016 IEEE International Conference on Pervasive Computing and
  Communication Workshops (PerCom Workshops)}, pages 1--6. IEEE, mar 2016.

\bibitem{Bloch2015}
Andreas Bloch, Robert Erdin, Sonja Meyer, Thomas Keller, and Alexandre
  de~Spindler.
\newblock {Battery-Efficient Transportation Mode Detection on Mobile Devices}.
\newblock In {\em 2015 16th IEEE International Conference on Mobile Data
  Management}, pages 185--190. IEEE, jun 2015.

\bibitem{Casale2011}
Pierluigi Casale, Oriol Pujol, and Petia Radeva.
\newblock {Human activity recognition from accelerometer data using a wearable
  device}.
\newblock {\em Pattern Recognition and Image Analysis}, 6669 LNCS:289--296,
  2011.

\bibitem{Hemminki2013}
Samuli Hemminki, Petteri Nurmi, and Sasu Tarkoma.
\newblock {Accelerometer-based transportation mode detection on smartphones}.
\newblock In {\em Proceedings of the 11th ACM Conference on Embedded Networked
  Sensor Systems - SenSys '13}, pages 1--14, New York, New York, USA, 2013. ACM
  Press.

\bibitem{hsu2003practical}
Chih-Wei Hsu, Chih-Chung Chang, Chih-Jen Lin, et~al.
\newblock A practical guide to support vector classification.
\newblock 2003.

\bibitem{Kang2015a}
Wonho Kang and Youngnam Han.
\newblock {SmartPDR: Smartphone-Based Pedestrian Dead Reckoning for Indoor
  Localization}.
\newblock {\em IEEE Sensors Journal}, 15(5):2906--2916, may 2015.

\bibitem{Krieg2016}
Jean~Gabriel Krieg, Gentian Jakllari, Hadrien Toma, and Andre~Luc Beylot.
\newblock {Unlocking the smartphone's senses for smart city parking}.
\newblock In {\em 2016 IEEE International Conference on Communications, ICC
  2016}, 2016.

\bibitem{Kwapisz2010}
Jennifer~R Kwapisz, Gary~M Weiss, and Samuel~A Moore.
\newblock {Activity Recognition using Cell Phone Accelerometers}.
\newblock {\em Human Factors}, 12(2):74--82, 2010.

\bibitem{Lara2013}
Oscar~D. Lara and Miguel~a. Labrador.
\newblock {A Survey on Human Activity Recognition using Wearable Sensors}.
\newblock {\em IEEE Communications Surveys {\&} Tutorials}, 15(3):1192--1209,
  2013.

\bibitem{Reddy2010}
Sasank Reddy, Min Mun, Jeff Burke, Deborah Estrin, Mark Hansen, and Mani
  Srivastava.
\newblock {Using mobile phones to determine transportation modes}.
\newblock {\em ACM Transactions on Sensor Networks}, 6(2):1--27, 2010.

\bibitem{Salpietro2016}
Rosario Salpietro, Luca Bedogni, Marco {Di Felice}, and Luciano Bononi.
\newblock {Park Here! a smart parking system based on smartphones' embedded
  sensors and short range Communication Technologies}.
\newblock In {\em IEEE World Forum on Internet of Things, WF-IoT 2015 -
  Proceedings}, pages 18--23, 2016.

\bibitem{Schilit1994}
Bill~N. Schilit, N.~Adams, and R.~Want.
\newblock {Context-aware computing applications}.
\newblock In {\em IEEE Workshop on Mobile Computing Systems and Applications},
  pages 85--90, 1994.

\bibitem{Stenneth2011}
Leon Stenneth, Ouri Wolfson, Philip~S. Yu, and Bo~Xu.
\newblock {Transportation mode detection using mobile phones and GIS
  information}.
\newblock {\em Proceedings of the 19th ACM SIGSPATIAL International Conference
  on Advances in Geographic Information Systems}, page~54, 2011.

\bibitem{Su2016}
Xing Su, Hernan Caceres, Hanghang Tong, and Qing He.
\newblock {Online Travel Mode Identification Using Smartphones with Battery
  Saving Considerations}.
\newblock {\em IEEE Transactions on Intelligent Transportation Systems},
  17(10):2921--2934, 2016.

\bibitem{zheng2010geolife}
Yu~Zheng, Xing Xie, and Wei-Ying Ma.
\newblock Geolife: A collaborative social networking service among user,
  location and trajectory.
\newblock {\em IEEE Data Eng. Bull.}, 33(2):32--39, 2010.

\end{thebibliography}
\end{document}